\documentclass[runningheads]{llncs}
\usepackage[square,numbers,sort&compress]{natbib}
\usepackage{acro}
\usepackage[textwidth=0.5in]{todonotes}
\usepackage{amsmath,amsfonts,amssymb}
\usepackage{hyperref}
\usepackage{graphicx}
\usepackage{multirow}
\usepackage{enumitem}
\usepackage{etoolbox}

\DeclareAcronym{sota}{
    long=state-of-the-art,
    short=SOTA,
}
\DeclareAcronym{dcp}{
    long=downstream classification performance,
    short=DCP,
}
\DeclareAcronym{stamp}{
    long=solid tumor associative modeling in pathology,
    short=STAMP
}
\DeclareAcronym{cnn}{
    long=convolutional neural network,
    short=CNN,
}
\DeclareAcronym{nlp}{
    long=natural language processing,
    short=NLP
}
\DeclareAcronym{vit}{
    long=Vision Transformer,
    short=ViT
}
\DeclareAcronym{ctp}{
    long=CTransPath,
    short=CTP
}
\DeclareAcronym{pvt}{
    long=Pyramid Vision Transformer,
    short=PVT
}
\DeclareAcronym{maxvit}{
    long=Multi-axis Vision Transformer,
    short=MaxViT
}
\DeclareAcronym{ma-sa}{
    long=Multi-axis Self-Attention,
    short=Max-SA
}
\DeclareAcronym{se}{
    long=Squeeze-and-Excitation,
    short=SE
}
\DeclareAcronym{mim}{
    long=masked image modeling,
    short=MiM
}
\DeclareAcronym{wsi}{
    long=whole slide image,
    short=WSI
}
\DeclareAcronym{mil}{
    long=multiple-instance learning,
    short=MIL
}
\DeclareAcronym{HE}{
    long=hematoxylin and eosin,
    short=H\&E
}
\DeclareAcronym{srcl}{
    long=semantically-relevant contrastive learning,
    short=SRCL
}
\DeclareAcronym{tcga}{
    long=The Cancer Genome Atlas,
    short=TCGA
}
\DeclareAcronym{paip}{
    long=Pathology AI Platform,
    short=PAIP
}
\DeclareAcronym{moco}{
    long=momentum contrast,
    short=MoCo
}
\DeclareAcronym{cptac}{
    long=Clinical Protemic Tumor Analysis Consortium,
    short=CPTAC
}
\DeclareAcronym{brca}{
    long=breast cancer,
    short=BRCA
}
\DeclareAcronym{vhio}{
    long=Vall d’Hebron Institute of Oncology,
    short=VHIO
}
\DeclareAcronym{n-sam}{
    long=negative sampling,
    short=N-Sam
}
\DeclareAcronym{ds}{
    long=dynamic sampling,
    short=DS
}
\DeclareAcronym{ops}{
    long=original positive sampling,
    short=OPS
}
\DeclareAcronym{mpp}{
    long=microns per pixel,
    short=mpp
}
\DeclareAcronym{wt}{
    long=wild-type,
    short=WT
}
\DeclareAcronym{mut}{
    long=mutated,
    short=MUT
}
\DeclareAcronym{la}{
    long=Luminal A,
    short=LA
}
\DeclareAcronym{lb}{
    long=Luminal B,
    short=LB
}
\DeclareAcronym{bl}{
    long=Basal-Like,
    short=BL
}
\DeclareAcronym{h2}{
    long=HER2 enriched,
    short=H2,
}
\DeclareAcronym{nl}{
    long=normal-Like,
    short=NL
}
\DeclareAcronym{auroc}{
    long=area under the receiver operating characteristic,
    short=AUROC
}
\DeclareAcronym{auprc}{
    long=area under the precision recall characteristic,
    short=AUPRC
}
\DeclareAcronym{ci}{
    long=confidence interval,
    short=CI
}
\DeclareAcronym{ssl}{
    long=self-supervised learning,
    short=SSL
}
\DeclareAcronym{ai}{
    long=Artificial Intelligence,
    short=AI
}
\DeclareAcronym{sra}{
    long=Spatial Reduction Attention,
    short=SRA
}
\DeclareAcronym{srq}{
    long=Sub-Research Question,
    short=SRQ
}
\DeclareAcronym{umap}{
    long=uniform manifold approximation and projection,
    short=UMAP
}

\begin{document}
\title{Reducing self-supervised learning complexity improves weakly-supervised classification performance in computational pathology}
\titlerunning{Reduced self-supervised learning complexity in computational pathology}
%
\author{Tim~Lenz\inst{1}
\and Omar~S.~M.~El Nahhas\inst{1}
\and Marta~Ligero\inst{1}
\and Jakob~Nikolas~Kather\inst{1,2,3}
}
\authorrunning{T. Lenz et al.}
%
\institute{Else Kroener Fresenius Center for Digital Health, Medical Faculty Carl Gustav Carus, TUD Dresden University of Technology, Germany \and
Department of Medicine 1, University Hospital and Faculty of Medicine Carl Gustav Carus, TUD Dresden University of Technology, Germany \and 
Medical Oncology, National Center for Tumor Diseases (NCT), University Hospital Heidelberg, Heidelberg, Germany
}
\maketitle              
\begin{abstract}
Deep Learning models have been successfully utilized to extract clinically actionable insights from routinely available histology data. Generally, these models require annotations performed by clinicians, which are scarce and costly to generate. The emergence of \ac{ssl} methods remove this barrier, allowing for large-scale analyses on non-annotated data. However, recent SSL approaches apply increasingly expansive model architectures and larger datasets, causing the rapid escalation of data volumes, hardware prerequisites, and overall expenses, limiting access to these resources to few institutions.
Therefore, we investigated the complexity of contrastive SSL in computational pathology in relation to classification performance with the utilization of consumer-grade hardware.
Specifically, we analyzed the effects of adaptations in data volume, architecture, and algorithms on downstream classification tasks, emphasizing their impact on computational resources.
We trained breast cancer foundation models on a large public patient cohort and validated them on various downstream classification tasks in a weakly supervised manner on two external public patient cohorts.
Our experiments demonstrate that we can improve downstream classification performance whilst reducing \ac{ssl} training duration by 90\%.
In summary, we propose a set of adaptations which enable the utilization of SSL in computational pathology in non-resource abundant environments. 
\end{abstract}

\begin{figure}[h!]
    \centering
    \includegraphics[width=0.45\paperwidth]{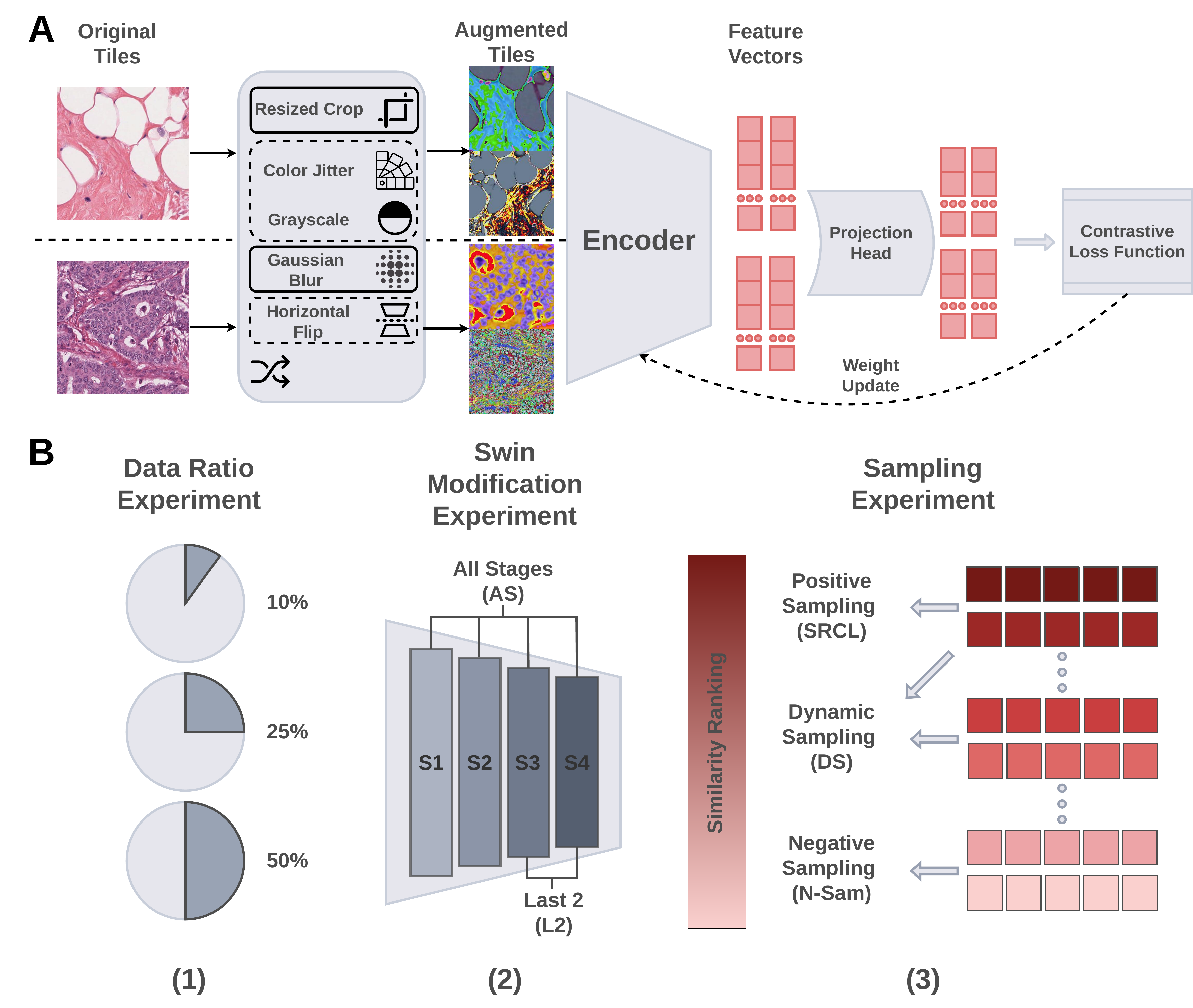}
    \caption{\textbf{Experiment Overview}. \textbf{(A)} visualizes the contrastive SSL algorithm. \textbf{(B)} shows an overview of the experiments we performed in this study. (1) We train on subsets of the \ac{ssl} data to analyze the impact of reduced data volumes during SSL on downstream classification performance (DCP). (2) We investigate the DCP of feature vectors retrieved from earlier stages of our SSL trained tiny Swin Transformer as well as combined feature representations of multiple layers. (3) We propose semantically relevant sampling methods to improve contrastive SSL in histopathology. Hence, for each feature representation of an input sample we rank all other representations of a batch based on the cosine similarity. Subsequently, we retrieve certain subsets of this ranking and adapt their weight in the contrastive loss function.}
    \label{fig:exp}
\end{figure}
\section{Introduction}
Artificial Intelligence (AI) tools can assist pathologists in various tasks like tumor detection~\cite{Wang2022}, disease classification~\cite{waqas2023revolutionizing} or prognosis~\cite{nahhas2023regression} and has even enabled the extraction of clinically actionable insights from routinely available data~\cite{unger2024systematic}. Traditionally, the development of AI models for medical applications has relied heavily on manually annotated data~\cite{spasic2020clinical}. 
Foundation models for computational pathology have proven to alleviate the need for annotated data as they are pre-trained on diverse datasets in order to be applied on various different use cases that are relevant for current research~\cite{wojcik2022foundation}.
In order to obtain adequate foundation models, massive amounts of data and compute are utilized~\cite{vorontsov2023virchow,Filiot2023MIMhisto,chen2023uni,campanella2023computational} to train large scale models like \acp{vit}~\cite{dosovitskiy2021}. The amount of training data can be substantially increased if the necessity of annotations is avoided. This led to the establishment of \ac{ssl} methods~\cite{campanella2023computational}. Instead of focusing on few targets during training, \ac{ssl} methods learn from inherent patterns and relationships in unlabeled data. Thereby, the capacity for generalization and adaptation to diverse tasks is amplified, while also surpassing the performance of models trained in a supervised manner~\cite{Kang2022,ding2023tailoring,krishnan2022self,chen2022selfsupervised}.
Foundation models trained with \ac{ssl} methods are often applied as encoders to compress the high-dimensional input data to its most important information content in the form of feature vectors. Those feature vectors are then applied on downstream classification tasks in a weakly supervised manner~\cite{unger2024systematic}.   
The rise of \ac{ssl} methods alongside exponential hardware improvements enabled the training of increasingly capable foundation models. 
Nevertheless, current strategies are too resource-intensive as the data volumes, hardware requirements and overall costs are increasing rapidly to a point where only few institutions can contribute to methodological improvements in this field of research. 
This point is exemplified in recent studies, which required thousands of GPU hours to train their foundation models in computational pathology~\cite{Filiot2023MIMhisto,Wang2022,chen2023uni,campanella2023computational,vorontsov2023virchow}.
This resource intensive entry-barrier limits open-source advancements in non-resource abundant environments.
Thus, we explore feasible solutions to train foundation models at scale on consumer-grade hardware. Our main contributions are as follows: 
\begin{itemize}[label=$\bullet$]
    \item To the best of our knowledge, we provide the first comprehensive ablation study that analyzes \ac{ssl} complexity reductions in relation to weakly supervised classification performance on breast cancer related histopathology tasks.
     \item We improve over existing semantically relevant contrastive learning methods in computational pathology.
    \item We show that modifications in data volume, the \ac{ssl} model architecure and the contrastive learning loss function  improve the efficiency of \ac{ssl} in computational pathology while surpassing \ac{sota} results. The findings of these adaptations enable the utilization of \ac{ssl} in computational pathology in non-resource abundant environments. 
\end{itemize}


\section{Methods}

We utilized \ac{moco}-v3 as our base contrastive \ac{ssl} framework. Following~\cite{Wang2022}, we chose the tiny Swin Transformer~\cite{liu2021swin} with a convolutional base as the default encoder architecture. In the following we will elucidate the approach and experimental settings for the \ac{ssl} encoder training and for the evaluation. An overview of the performed experiments can be found in \autoref{fig:exp}.

\subsection{Contrastive Self-Supervised Learning}
For Histopathology tasks, contrastive methods exhibit a decent compromise between performance, simplicity and efficiency~\cite{Kang2022,stacke2022learning}.
Therefore, we employ the third version of \ac{moco}~\cite{Chen2021mocoV3} as the basis for our \ac{ssl} experiments (\autoref{fig:exp}-A). Following He et al.~\cite{He2019moco} we interpret contrastive learning as training an encoder for a \textit{dictionary look-up task}: \\
Let a set of encoded samples $\{k_0,k_1,k_2,\dots,k_N\}$ be the keys of a dictionary, such that exactly one matching key $k^+$ exists for a query $q$. Then, the contrastive loss is low if $q$ is similar to $k^+$ and dissimilar to all other keys. The InfoNCE~\cite{Oord2019} loss function is then defined as: 
\begin{equation}\label{eq:moco}
\mathcal{L}_{\mathbf{q}} = -\log \frac{\psi(\mathbf{q},\mathbf{k^+})}{\psi(\mathbf{q},\mathbf{k^+})+\sum_{i=1}^K\psi(\mathbf{q},\mathbf{k_i})}, 
\end{equation}
where $q$ and the corresponding $k^+$ are the feature vectors belonging to different random augmentations of the same input image and $K$ is the number of negative samples which is equal to $N-1$, with $N$ being the batch size or length of the memory queue.
The function $\psi$ is defined as follows:
\begin{equation}\label{eq:psi}
    \psi(\mathbf{x_1},\mathbf{x_2}) = \exp(\text{sim}(\mathbf{x_1},\mathbf{x_2})/\tau),
\end{equation}
where $\tau$ denotes the temperature parameter and the cosine similarity function is denoted as $\text{sim}(\cdot)$.
To avoid feature collapse, the keys and queries need to be generated by distinct encoders. Let $\theta_q$ denote the parameters of the query encoder with the dense projection head, then the parameters of the key encoder $\theta_k$ are updated as follows: 
\begin{equation}\label{eq:ke}
    \theta_k\leftarrow m\theta_k+(1-m)\theta_q,
\end{equation}
where $m\in [0,1)$ is the momentum coefficient. With the key encoder as the exponential average of the query encoder, the key representations stay more consistent which enables a more stabilized training process. 
We adapted the public MoCo-v3 repository\footnote{\url{https://github.com/facebookresearch/moco-v3}} for our experiments to train the tiny Swin Transformer encoders from scratch over 50 epochs using the AdamW~\cite{loshchilov2019} optimizer, a learning rate of $1.5\cdot 10^{-4}$, a batch size of $1024$ and 40 warm-up epochs. Besides the batch size and the total number of epochs, all other settings remain the default options stated in the documentation of the github repository. 

\subsection{Semantically Relevant Sampling}
Wang et al.~\cite{Wang2022} proposed a self-supervised contrastive learning framework for histopathology called \textit{\ac{srcl}}. 
The main difference between \ac{moco} and \ac{srcl} is the increased number of positive samples~\cite{Wang2022,Chen2021mocoV3}. 
Increasing the number of positive samples is of special interest for histopathology tasks as tessellated patches of the same \ac{wsi} can exhibit high similarity and should therefore not necessarily be regarded as negative samples during contrastive learning. 
\subsubsection{Positive Sampling.}
Positive sampling is based on the similarity of the encoded patches and independent on their spatial location in the \ac{wsi}.
Specifically, we sample $S=5$ additional positives based on the highest cosine similarity between the key encoder output of an original sample and the other key encoder representations of the augmented samples belonging to a batch or memory queue (cf. \autoref{eq:moco}):
\begin{equation}\label{eq:ps}
\mathcal{L}_{\mathbf{q}} = -\log \frac{\sum_{j=1}^{S+1}\psi(\mathbf{q},\mathbf{k_j^+})}{\sum_{j=1}^{S+1}\psi(\mathbf{q},\mathbf{k_j^+})+\sum_{i=1}^K\psi(\mathbf{q},\mathbf{k_i})}, 
\end{equation}
where $K=N-S-1$ and $\psi$ defined in \autoref{eq:psi}.
Following ~\cite{Wang2022} we start the sampling of additional positives after 5 epochs of traditional contrastive learning as the representations need to be meaningful enough such that similar representations relate to similar input patches. We will refer to positive sampling as \ac{srcl} following~\cite{Wang2022}.

\subsubsection{Negative Sampling.}
Due to the success of positive sampling proposed by Wang et al.~\cite{Wang2022}, we decided to further explore the effects of semantically relevant sampling on the performance of histopathology encoders. 
We retrieve the $T=50$ most dissimilar representations in a batch for each sample. However, instead of ranking the similarity of representations of original patches with representations of augmented patches from the same batch, we rank the similarity of the representations of original patches processed by the key encoder. It is important to observe that
this even reduces the computational requirements as you only have to sample one set of
logits while in \ac{srcl} you have to do it separately for the two different augmented views.
The loss function for \ac{n-sam} is updated as follows:
\begin{equation}\label{eq:ns}
\mathcal{L}_{\mathbf{q}} = -\log \frac{\psi(\mathbf{q},\mathbf{k^+})}{\psi(\mathbf{q},\mathbf{k^+})+\sum_{j=1}^T\psi(\mathbf{q},\mathbf{k_j^-})+\sum_{i=1}^K\psi(\mathbf{q},\mathbf{k_i})},
\end{equation}
with $K=N-1$ and $\psi$ defined in \autoref{eq:psi}.

\subsubsection{Dynamic Sampling.}
Because contrasting most similar input patches is unnecessary and distinguishing highly dissimilar samples is trivial, we hypothesize that increasing the weights for the samples of intermediate similarity would be beneficial. 
Therefore, we propose \ac{ds}, where we increase the number of negative samples by 20 every epoch $>5$ and decrease the number of positive sample from 30 to 1 in increments of 5 every epoch $>5$. Furthermore, instead of sampling negative samples from the bottom of the similarity ranking, we sample from the middle of the ranking growing towards both sides. The loss function for DS is as follows:
\begin{equation}\label{eq:ds}
\mathcal{L}_{\mathbf{q}} = -\log \frac{\sum_{j=1}^{S_e+1}\psi(\mathbf{q},\mathbf{k_j^+})}{\sum_{j=1}^{S_e+1}\psi(\mathbf{q},\mathbf{k_j^+})+\sum_{j=1}^{T_e}\psi(\mathbf{q},\mathbf{k_i^-})+\sum_{i=1}^K\psi(\mathbf{q},\mathbf{k_i})},
\end{equation}
where $K=N-S-1$ and $\psi$ defined in \autoref{eq:psi}, the $e$ in $S_e$ and $T_e$ indicates their dependency on the epoch.

\subsection{Swin Transformer Modifications}

We removed Swin Transformer~\cite{liu2021swin} stages after \ac{ssl} to extract features from earlier stages and evaluate their informational content on downstream classification tasks.
Since the tensor shapes of earlier stages differ from the output dimension we applied 2D adaptive average pooling to retrieve 768 features per patch.
Additionally, we concatenated features from all four stages as well as from the last two stages to analyze their combined value for downstream classification tasks.

\subsection{Evaluation}

We follow the STAMP protocol~\cite{nahhas2023wholeslide} for our data processing pipeline to evaluate the \ac{ssl} encoders.
It consists of two major steps: \ac{wsi} preprocessing and weakly-supervised training. 
We resize the images to 0.5~\ac{mpp}, use canny edge detection~\cite{rong2014canny} to reject the background and tessellate the tissue part into $n$ patches of size $224\times 224$ pixels. 
These patches are then processed by an encoder to compress their information to feature vectors of dimension 768. Those embeddings are the result of the \ac{wsi}-pre-processing and are saved as intermediate files. Next, the weakly supervised training uses those embeddings as inputs to predict a \ac{wsi}-level categorical target. A \ac{vit} model~\cite{dosovitskiy2021} with two layers and 8 attention heads each is used for the supervised \ac{mil} training following Wagner et al.~\cite{wagner2023transformer}. 
Since we focus on breast cancer tissue samples, we chose targets with clinical relevance as they are important for prognostic implications. 
Among the most commonly altered genes with prognostic importance in breast cancer are \textit{TP53}, \textit{CDH1} and \textit{PIK3CA}~\cite{shiovitz2015genetics,andre2021alpelisibPik3ca,li2006pik3ca}. Therefore, we evaluate the \ac{ssl} encoders on the mutation information of those genes. 
To increase the diversity of our targets even further, we also evaluate our models on metastasis detection in lymph nodes~\cite{bejnordi2017camelyon}.
We use \ac{tcga}-\ac{brca} as the training cohort for the genes \textit{TP53}, \textit{CDH1} and \textit{PIK3CA}. 
The publicly available CAMELYON16 challenge~\cite{bejnordi2017camelyon} provides internal and external cohorts with the necessary target information about metastasis detection.
We use the breast cancer subset of the \ac{cptac} as external cohorts for the targets \textit{TP53}, \textit{CDH1} and \textit{PIK3CA}.
The \ac{vit} \ac{mil} models are trained in a weakly-supervised manner on the internal cohort with five-fold cross-validation. Subsequently, the models are deployed on the external cohorts.
We report \ac{auprc} (\autoref{tab:results-prc}) and \ac{auroc} scores (\autoref{tab:results-roc}) as evaluation metrics, with \ac{auprc} being the main metric due to the data imbalance~\cite{wagner2023transformer}.

\section{Experimental Results and Discussion} %
\subsubsection{Data.}
We use public datasets throughout this study. For the SSL training, we utilize the breast cancer (BRCA) subset of \ac{tcga} (\url{www.cbioportal.org}) consisting of 1125 \ac{HE} stained \acp{wsi}. The \acp{wsi} are tessellated into patches with 0.5~\ac{mpp} of size $512 \times 512$ pixels. After rejecting the background, we retrieve over 4M patches for the \ac{ssl} training. For the downstream evaluation tasks we employ TCGA-BRCA, the BRCA subset of \ac{cptac} (\url{https://proteomics.cancer.gov/programs/cptac}) and tissue slides from the CAMELYON16 challenge~\cite{bejnordi2017camelyon}. An overview of the utilized datasets can be found in \autoref{tab:data}. The performed experiments are visualized in \autoref{fig:exp}-B.
\begin{table}[h]
    \centering
    
\begin{tabular}{l|c|c|c|c|c}
& CPTAC-BRCA & CPTAC-BRCA & CPTAC-BRCA & CAMELYON16 & \\
MODEL & \textit{CDH1} & \textit{TP53}  & \textit{PIK3CA} & TUMOR & AVG \\
\hline
\hline
\textbf{CTransPath} & \textbf{0.991} & 0.793 & 0.760 & 0.683 & 0.807 \\
\hline
MoCo-v3& 0.984 & 0.823 & 0.790 & 0.415 & 0.753 \\
\hline
50\% data & 0.987 & 0.807 & 0.795 & 0.422 & 0.753 \\
25\% data & 0.981 & 0.739 & 0.764 & 0.627 & 0.778 \\
10\% data & 0.983 & 0.706 & 0.751 & 0.557 & 0.749 \\
\hline
Stage 3 & 0.973 & 0.808 & 0.760 & 0.759 & 0.825 \\
Stage 2 & 0.961 & 0.686 & 0.772 & 0.476 & 0.724 \\
Stage 1 & 0.958 & 0.695 & 0.726 & 0.455 & 0.709 \\
All Stages & 0.968 & 0.798 & 0.762 & 0.794 & 0.830 \\
Last 2 & 0.972 & \textbf{0.831} & 0.745 & \textbf{0.811} & \textbf{0.840} \\
\hline
SRCL & 0.986 & 0.802 & 0.778 & 0.416 & 0.746 \\
N-Sam & 0.983 & 0.821 & \textbf{0.795} & 0.436 & 0.759 \\
DS & 0.989 & 0.783 & 0.791 & 0.457 & 0.755 \\
\end{tabular}
        \caption{Comparison of the encoders deployed on downstream classification tasks. The table contains the average Area Under the Precision Recall Characteristic (AUPRC) over all five Vision Transformer models from the cross validation. SRCL: semantically relevant contrastive learning, N-Sam: negative sampling, DS: dynamic sampling.}\label{tab:results-prc}
\end{table}
\subsubsection{Our \ac{moco} encoder improves over \ac{sota} results on breast cancer gene classification tasks.}
We trained a tiny Swin transformer with \ac{moco}-v3 over half the number of epochs and on only 4\% of the \acp{wsi} compared to the \ac{sota} \ac{ctp} model, and it reaches similar performance on the downstream gene classification tasks (\autoref{tab:results-prc}). 
On the gene classification targets \textit{TP53} and \textit{PIK3CA}, our model exhibits a superior performance compared to \ac{ctp} ($\text{AUPRC}_\text{\textit{TP53}}=79.3\%;~\text{AUPRC}_\text{\textit{PIK3CA}}=76.0\%$) as the results improve $+3\%$ in \ac{auprc} for both targets. However, it exhibits substantially worse performance on the metastasis detection task from the CAMELYON16 test set. 
This drop in performance is presumed to be due to the amplified number of tissue types \ac{ctp} has been trained on as \ac{tcga} even includes lymph node \acp{wsi} from axial regions. This is exactly the tissue type utilized for the CAMELYON16 challenge. 

\subsubsection{A 50\% reduction in SSL training data does not affect downstream gene mutation prediction performance.}
With the data reduction experiment we aim to investigate how limiting \ac{ssl} samples impacts the \ac{dcp} of them. 
The potential for the reduction of computational costs is substantial while differences in performance might be negligible~\cite{chen2023uni}. 
The results show that 50\% of the \ac{ssl} data is adequate to achieve equivalent downstream performance as the 100\% baseline (75.3\% average AUPRC). Therefore, we can achieve a 50\% reduction of computational resources and time without any loss in classification capabilities.
\subsubsection{Incorporating feature representations from previous blocks improves weakly supervised learning performance.}
The representations extracted from the first two stages display inferior performance across all targets.
However, our findings demonstrate strongly improved performance on the metastasis detection task for all the models that use the features extracted from the third stage of the encoder (\acp{auprc}: CTP: 68.3\%, MoCo: 41.5\%, S3: 75.9\%, AS: 79.4\%, L2: 81.1\%). Although, these encoders exhibit slightly reduced \ac{auprc} scores on \textit{CDH1} compared to \ac{ctp} and our \ac{moco} baseline, they improve over \ac{ctp} on \textit{TP53}. Consequently, depending on the task at hand, removing the last stage of the encoder for the feature extraction not only reduces computational costs but also improves classification performance.   

\subsubsection{Negative sampling and dynamic sampling improve over semantically relevant contrastive learning.}

Our proposed methods \ac{ds} and \ac{n-sam} exhibit the best performance on \textit{PIK3CA} (AUPRC=79.1\%, 97.5\%) and improve over the \ac{moco} baseline (+2.1\%,+4.2\% AUPRC) and \ac{srcl} (+2\%,+4.1\% AUPRC) on the metastasis detection task. However, contrary to previous results~\cite{Wang2022}, \ac{srcl} does not improve over the \ac{moco}-v3 baseline for our tasks. This could be due to the difference in the tasks at hand. While we evaluate the \ac{ssl} algorithms utility with weakly supervised \ac{mil} tasks, Wang et al.\ evaluated the benefit of \ac{srcl} on a tile based supervised task~\cite{Wang2022}.

\section{Conclusion}

In this work, we analyzed potential avenues to reduce the required resources to train breast cancer \ac{wsi} encoders in a contrastive self-supervised manner while improving the classification performance on gene mutation and metastasis detection tasks. 
Our experimental framework enables the benchmarking of a variety of modifications regarding the contrastive \ac{ssl} based on \ac{moco}-v3 in computational pathology. We examined the influence of data volume adaptations, architecture changes and algorithmic adjustments to analyze their effect on downstream classification tasks while reducing computational complexity.
Our findings show that \ac{dcp} was not effected despite a 50\% reduction of \ac{ssl} data. 
Furthermore, we demonstrate that removing the last stage of the encoder can enhance performance of weakly supervised learning tasks. Therefore, by reducing the number of parameters, we improve current approaches for feature extraction while making them less computationally expensive.
In summary, we provide the methodological insights to enable foundation model training on routinely available data for medical centers across the globe.
\bibliographystyle{splncs04} 
\bibliography{literature.bib}

\newpage
\appendix 
\title{Reduced self-supervised learning complexity improves weakly-supervised classification performance in computational pathology}
\titlerunning{Reduced self-supervised learning complexity in computational pathology}
%
\author{Tim~Lenz\inst{1}\orcidID{0000-0002-9034-2535} \and
Omar~S.M.~El Nahhas\inst{1}\orcidID{0000-0002-2542-2117} \and
Marta~Ligero\inst{1}\orcidID{0000-0001-9824-7316}
\and Jakob~Nikolas~Kather\inst{1,2,3,4}\orcidID{0000-0002-3730-5348}
}
\authorrunning{T. Lenz et al.}
%
\institute{Else Kroener Fresenius Center for Digital Health, Medical Faculty Carl Gustav Carus, TUD Dresden University of Technology, Germany \and
Department of Medicine 1, University Hospital and Faculty of Medicine Carl Gustav Carus, TUD Dresden University of Technology, Germany \and 
Pathology \& Data Analytics, Leeds Institute of Medical Research at St James’s, University of Leeds, Leeds, United Kingdom \and
Medical Oncology, National Center for Tumor Diseases (NCT), University Hospital Heidelberg, Heidelberg, Germany
}

\section{Supplementary Material}
\begin{table}[h]
\centering
\begin{tabular}{c|c|c|ccc|lr|lr}
\textbf{Target} & \textbf{Cohort} & \textbf{i. / e.} & \textbf{\#WSI} & \#feats & \textbf{\#t.o.} & \textbf{C1} & \textbf{\#} & \textbf{C2} & \textbf{\#} \\
\hline
\textit{CDH1} & TCGA-BRCA & internal & 1133 & 1125 & 1022 & WT & 900 & MUT & 122 \\

\textit{TP53} & TCGA-BRCA & internal & 1133 & 1125 & 1022 & WT & 683 & MUT & 339 \\

\textit{PIK3CA} & TCGA-BRCA & internal & 1133 & 1125 & 1022 & WT & 686 & MUT & 336 \\

tumor & CAMELYON16 & internal & 270 & 265 & 265 & NO & 157 & TU & 108  \\
\hline
\textit{CDH1} & CPTAC-BRCA & external & 653 & 653 & 120 & WT & 112 & MUT & 8 \\

\textit{TP53} & CPTAC-BRCA & external & 653 & 653 & 120 & WT & 71 & MUT & 49 \\

\textit{PIK3CA} & CPTAC-BRCA & external & 653 & 653 & 120 & WT & 82 & MUT & 38\\
tumor & CAMELYON16 & external & 129 & 129 & 129 & NO & 80 & TU & 49 \\
\end{tabular}
\caption{Overview of the targets used to evaluate the capabilities of the trained encoders. Abbreviations: i. / e. - internal (weakly supervised training) or external (deployment), feats - features (slides processed by the encoder), t.o. - target overlap (intersection between processed slides and available target information), C1 \& C2 - available classes, WT - wild-type, MUT - mutated, NO - normal, TU - tumor.}
\label{tab:data}
\end{table}
\begin{table}[h!]
    \centering
\begin{tabular}{l|c|c|c|c|c}
& CPTAC-BRCA & CPTAC-BRCA & CPTAC-BRCA & CAMELYON16 &  \\
MODEL & \textit{CDH1} & \textit{TP53}  & \textit{PIK3CA} & TUMOR & AVG \\
\hline
\hline
\textbf{CTransPath} & \textbf{0.894} & 0.720 & 0.596 & 0.686 & \textbf{0.724} \\
\hline
MoCo-v3 & 0.842 & \textbf{0.750} & 0.647 & 0.519 & 0.689 \\
\hline
50\% data & 0.861 & 0.729 & 0.648 & 0.500 & 0.685 \\
25\% data & 0.826 & 0.646 & 0.613 & 0.623 & 0.677 \\
10\% data & 0.799 & 0.612 & 0.585 & 0.556 & 0.638 \\
\hline
Stage 3 (S3) & 0.759 & 0.713 & 0.601 & 0.756 & 0.708 \\
Stage 2 (S2) & 0.632 & 0.552 & 0.611 & 0.528 & 0.581 \\
Stage 1 (S1) & 0.579 & 0.556 & 0.545 & 0.531 & 0.553 \\
AS & 0.729 & 0.716 & 0.601 & 0.779 & 0.706 \\
L2 & 0.751 & 0.750 & 0.590 & \textbf{0.796} & 0.722 \\
\hline
SRCL & 0.846 & 0.720 & 0.624 & 0.545 & 0.684 \\
N-Sam & 0.832 & 0.738 & 0.647 & 0.568 & 0.696 \\
DS & 0.871 & 0.708 & \textbf{0.654} & 0.554 & 0.697 \\
\end{tabular}
    \caption{External \ac{auroc} Results. Comparison of the encoders deployed on downstream classification tasks. The table contains the average \ac{auroc} over all five \ac{vit} models from the cross validation. }
    \label{tab:results-roc}
\end{table}

\end{document}